\documentclass[preprint]{article}

\usepackage{neurips_2023}

\usepackage[utf8]{inputenc} % allow utf-8 input
\usepackage[T1]{fontenc}    % use 8-bit T1 fonts
\usepackage{hyperref}       % hyperlinks
\usepackage{url}            % simple URL typesetting
\usepackage{booktabs}       % professional-quality tables
\usepackage{amsfonts}       % blackboard math symbols
\usepackage{nicefrac}       % compact symbols for 1/2, etc.
\usepackage{microtype}      % microtypography
\usepackage{xcolor}         % colors
\usepackage{listings}
\usepackage{cite}
\usepackage{graphicx}
\usepackage{hyperref}
\usepackage{array}
\usepackage{flafter}
\usepackage[export]{adjustbox} % Required for valign option

\hypersetup{
    pdftitle={MMMModal - Multi-Images Multi-Audio Multi-turn Multi-Modal},
    pdfauthor={Husein Zolkepli, Aisyah Razak,  Kamarul Adha, Ariff Nazhan},
    pdfsubject={Natural Language Processing, Large Language Models, Machine Learning},
    pdfkeywords={language models, natural language processing, deep learning},
    colorlinks=true,
    linkcolor=blue,
    citecolor=blue,
    urlcolor=blue
}
\title{MMMModal - Multi-Images Multi-Audio Multi-turn Multi-Modal}
\author{
  Husein Zolkepli\thanks{husein@mesolitica.com} \and
  Aisyah Razak\thanks{aisyahrazak171@gmail.com} \and
  Kamarul Adha\thanks{kamarul.adha360@gmail.com} \and
  Ariff Nazhan\thanks{ariffnzhn@gmail.com}
}

\begin{document}

\maketitle

\begin{abstract}

  Our contribution introduces a groundbreaking multimodal large language model designed to comprehend multi-images, multi-audio, and multi-images-multi-audio within a single multiturn session. Leveraging state-of-the-art models, we utilize the SigLIP encoder for visual inputs and the Whisper Encoder for audio inputs. Notably, this multimodal large language model is bilingual, proficient in understanding both English and Malay simultaneously. We proudly unveil two versions of this model: TinyLlama with 1.1B parameters, and Mistral with 7B parameters. With its ability to navigate diverse modalities and languages, our model represents a significant advancement for the Malaysian context and beyond.

  All models released at \href{https://huggingface.co/collections/mesolitica/multimodal-malaysian-llm-65c6f893e03f78fa9e5c8859}{HuggingFace Mesolitica Multimodal Malaysian LLM}.

\end{abstract}

\section{Introduction}

Language models trained with instructions have demonstrated remarkable performance across various domains. However, their limitation in handling only text-based data hampers their applicability. Recent advancements in multimodal pre-training have shown the potential to integrate knowledge from diverse modalities into a unified representation \cite{lyu2023macawllm,liu2023visual,openai2023gpt4}.

The introduction of OpenAI's GPT-4 \cite{openai2023gpt4}, which incorporates LLM with visual understanding capability, marked a milestone in the industry's progress, demonstrating significant advancements in addressing open-ended visual question-answering (VQA) tasks. Pioneering research initiatives such as LLaVA \cite{liu2023visual} and MiniGPT-4 \cite{zhu2023minigpt4} provide insightful directions in visual and text understanding capability. Their findings suggest that by incorporating visual encoders into existing LLMs and fine-tuning them using multi-modal instruction-tuning datasets, LLMs can be effectively transformed into multimodal LLMs. While Macaw-LLM \cite{lyu2023macawllm} introduces the integration of LLM with four different modalities: text, audio, video, and images. They successfully process information from different inputs effectively, enabling seamless information retrieval across domains. Existing dataset for multimodal instruction made available from \cite{liu2023visual} primarily supports instruction-following data involving visual content for conversation, detailed description and complex reasoning data.

Despite recent advancements, there remains a lack of current research on multimodal models capable of handling multiple images or audio inputs along with multi-turn dialogue. Moreover, there is a lack of existing multimodal datasets incorporating multi-turn interactions with multiple audio and image inputs, and little consideration has been given to the Malaysian context. To address these gaps, our proposal introduces MMMModal, a multimodal large language model fine-tuned for multi-modal instruction, integrating multiple image, audio, and text modalities within a single model architecture. Additionally, we present a corpus and employ an adaptive synthetic data generation method tailored to provide access to multi-image, multi-audio, multi-turn datasets with regards to languages in Malay and English.

\begin{itemize}

  \item \textbf{Synthetic Audio Instruction Dataset:} To construct Synthetic Audio Instruction Dataset, our approach involves gathering extracted audio content from YouTube videos. We  employed the Whisper Large V3 model for pseudolabeling to transcribe the audio contents from scraped Youtube videos, followed by a post-filtering process based on score thresholds to ensure high-quality datasets. We then utilized the Mixtral-8x7B-Instruct-v0.1 Model to generate multiturn dialogue instruction-following data involving the audio context.

  \item \textbf{Synthetic Visual Malaysian Context Dataset:}  We collected images from Malaysian websites along with their descriptions. Through data processing, we converted this information into conversational examples to better incorporate Malaysian context into our model.

  \item \textbf{Synthetic Multi-Images Multi-Audio relationship Dataset:} Our method involved randomly combining images, audios or pairing them together to create a dataset illustrating relationships of different images and audios. We utilized Mistral to generate multiturn dialogues, prompting the model to respond based on the images, audio captions, and descriptions. This enabled us to collect a corpus tailored for focusing on relationships within multi-modal content, encompassing both images and audio.

  \item \textbf{Pretraining Feature Alignment:} Our approach adopts a two-step training procedure to integrate multimodal and multiturn capabilities into our model. The initial step entails pretraining the feature alignment module. Through this process, we align the image and audio features with the pre-trained word embeddings of the Large Language Model (LLM). Specifically, this step involves training the projection layer to ensure alignment between the multi-modal features and textual representations. This alignment facilitates seamless integration of diverse modalities within the model architecture.

  \item \textbf{Finetuned Multi-Images Multi-Audio Multi-turn Model:} After pre-training for feature alignment, the projector module becomes familiar with the visual and embedding space. However, it still lacks the capability to discern the finer details of images and audios, or to respond to human questions and instructions effectively. In the second stage, we utilize generated synthetic Multimodal data to enhance performance and further align the embeddings with the Large Language Model (LLM). This step improves the LLM's ability to produce more natural and reliable language outputs for multimodal instructions.

\end{itemize}

\section{Synthetic Data Generation for Audio Instructions}\label{sec:synthetic-audio}

In our pursuit of creating a comprehensive dataset reflective of the Malaysian and Singaporean digital landscape, we meticulously compiled an extensive corpus comprising over 18,500 hours of YouTube content. This rich repository spans a diverse spectrum of subjects, ranging from the intricacies of gaming and the dynamics of economics to the nuances of political discourse, religious insights, cinematic endeavors, and the multifaceted dimensions of socioeconomics. Leveraging advanced AI technologies, specifically OpenAI's Whisper Large V3 model \cite{radford2022whisper}, we embarked on the ambitious task of pseudolabeling the transcriptions of these videos. This process enabled us to extract textual representations from the audiovisual content, laying a robust foundation for subsequent endeavors in generating multi-turn conversations based on the context embedded within these transcriptions.

Utilizing the transcriptions as contextual prompts, we embarked on the process of simulating dialogue interactions between users and virtual assistants. Leveraging the Mixtral-8x7B-Instruct-v0.1 model \cite{jiang2024mixtral}, specifically tailored for generating conversational responses, we orchestrated dynamic exchanges that mirrored real-world interactions. Through this approach, we sought to capture the nuances of human conversation, including the fluidity of dialogue and the contextual relevance of responses. By intertwining the audio context derived from the YouTube video transcriptions with the conversational capabilities of the model.

Furthermore, to ensure linguistic fidelity and cultural resonance, we employed Neural Machine Translation to post-translate the generated output from the Mixtral-8x7B-Instruct-v0.1 model into the Malay language, thereby enriching the dataset with linguistic diversity and contextual relevance tailored for the Malaysian audience.

In the first step, we create questions from the provided context. This process involves carefully analyzing the context and using advanced techniques to generate relevant questions. These questions aim to capture the key points and details of the context effectively. Below is the prompt we use to generate the questions,

\begin{lstlisting}[breaklines=true]
  paragraph = 'anda tahu keuntungan boleh lebih tinggi daripada keuntungan kewangan ...'
  instruction = f'{paragraph}\n\ngenerate questions based on context above'
  mixtral(instruction)
\end{lstlisting}

In the second step, we repeatedly refine and expand upon the initial questions. This iterative process helps us cover various aspects of the context more thoroughly, resulting in a broader range of questions, the pseudo Python code to generate the synthetic multo-turn as below,

\begin{lstlisting}[breaklines=true]
  def format_prompt(message, history):
    prompt = "<s>"
    for user_prompt, bot_response in history:
        prompt += f"[INST] {user_prompt} [/INST]"
        prompt += f" {bot_response}</s> "
    prompt += f"[INST] {message} [/INST]"
    return prompt

  def format_user(history):
    prompt = "<s>"
    for user_prompt, bot_response in history:
        prompt += f"[INST] {user_prompt} [/INST]"
        prompt += f" {bot_response}</s> "
    prompt += f"[INST]"
    return prompt
  
  paragraph = 'anda tahu keuntungan boleh lebih tinggi daripada keuntungan kewangan ...'
  initial_question = 'Why might someone consider investing in cryptocurrencies like Ethereum instead of traditional financial investments such as real estate?'
  prompt = f'{paragraph}\n{initial_question}'
  formatted_prompt = format_prompt(question, [])
  answer = mixtral(formatted_prompt)

  history = [(prompt, answer)]
  for _ i in range(N):
    formatted_prompt = format_user(history)
    question = mixtral(formatted_prompt)

    formatted_prompt = format_prompt(question, history)
    answer = mixtral(formatted_prompt)

    history.append((question, answer))

\end{lstlisting}

An example of generated dataset,

\begin{lstlisting}[breaklines=true]
{'context': 'anda tahu keuntungan boleh lebih tinggi daripada keuntungan kewangan rumah maka saya tidak akan mencari dalam akaun saya akan mencari ke dalam ethereum atau beberapa crypto punks bergantung pada faktor risiko anda kerana rumah kajang dihantar tidak mengganggu dsr saya sejauh ini jadi sekarang apa posisi saya untuk mendapatkan kewangan ketiga jadi mungkin setelah melihat sekeliling saya menemui seorang penjual yang dapat menutupi perhubungan tetapi bank hanya menerima 70% dari itu saya boleh membayar perbezaan dengan menggunakan wang ini kerana sekali lagi ia menyusahkan saya dan aset tetapi jika anda tidak selesa dengan mencari',
'chat': [{'role': 'user',
  'content': 'Why might someone consider investing in cryptocurrencies like Ethereum instead of traditional financial investments such as real estate?',
  'content_ms': 'Mengapakah seseorang mungkin mempertimbangkan untuk melabur dalam mata wang kripto seperti Ethereum dan bukannya pelaburan kewangan tradisional seperti hartanah?'},
  {'role': 'assistant',
  'content': 'There are several reasons why someone might consider investing in cryptocurrencies like Ethereum instead of traditional financial investments such as real estate:\n\n1. Potential for higher returns: Cryptocurrencies have the potential for much higher returns compared to traditional financial investments. While the real estate market can be profitable, the potential for exponential growth in the cryptocurrency market is significantly higher.\n2. Diversification: Investing in cryptocurrencies can provide a diversification benefit to an investment portfolio. By investing in a variety of assets, including cryptocurrencies, investors can spread their risk and potentially increase their overall returns.\n3. Liquidity: Cryptocurrencies are generally more liquid than real estate investments. It is easier to buy and sell cryptocurrencies, and they can be quickly converted into cash.\n4. Accessibility: Investing in cryptocurrencies is generally more accessible than investing in real estate. There are fewer barriers to entry, and it is easier to get started with a smaller amount of money.\n5. Decentralization: Cryptocurrencies are decentralized, meaning they are not controlled by any government or financial institution. This can provide a level of independence and autonomy that is not possible with traditional financial investments.\n\nHowever, it is important to note that investing in cryptocurrencies is also riskier than traditional financial investments. The value of cryptocurrencies can be highly volatile, and there is a risk of losing all or a significant portion of the investment. Therefore, it is important to carefully consider your own risk tolerance and financial situation before investing in cryptocurrencies.',
  'content_ms': 'Terdapat beberapa sebab mengapa seseorang mungkin mempertimbangkan untuk melabur dalam mata wang kripto seperti Ethereum dan bukannya pelaburan kewangan tradisional seperti hartanah:\n\n1. Potensi untuk pulangan yang lebih tinggi: Mata wang kripto mempunyai potensi untuk pulangan yang lebih tinggi berbanding dengan pelaburan kewangan tradisional. Walaupun pasaran hartanah boleh menguntungkan, potensi pertumbuhan eksponen dalam pasaran mata wang kripto adalah jauh lebih tinggi.\n2. Kepelbagaian: Melabur dalam mata wang kripto boleh memberikan manfaat kepelbagaian kepada portfolio pelaburan. Dengan melabur dalam pelbagai aset, termasuk mata wang kripto, pelabur boleh menyebarkan risiko mereka dan berpotensi meningkatkan pulangan keseluruhan mereka.\n3. Kecairan: Mata wang kripto biasanya lebih cair daripada pelaburan hartanah. Lebih mudah untuk membeli dan menjual mata wang kripto, dan ia boleh ditukar dengan cepat kepada wang tunai.\n4. Kebolehcapaian: Melabur dalam mata wang kripto secara umumnya lebih mudah diakses daripada melabur dalam hartanah. Terdapat lebih sedikit halangan untuk masuk, dan lebih mudah untuk bermula dengan jumlah wang yang lebih kecil.\n5. Desentralisasi: Mata wang kripto adalah terdesentralisasi, bermakna ia tidak dikawal oleh mana-mana kerajaan atau institusi kewangan. Ini boleh memberikan tahap kebebasan dan autonomi yang tidak mungkin dilakukan dengan pelaburan kewangan tradisional.\n\nWalau bagaimanapun, adalah penting untuk ambil perhatian bahawa melabur dalam mata wang kripto juga lebih berisiko daripada pelaburan kewangan tradisional. Nilai mata wang kripto boleh menjadi sangat tidak menentu, dan terdapat risiko kehilangan semua atau sebahagian besar pelaburan. Oleh itu, adalah penting untuk mempertimbangkan dengan teliti toleransi risiko dan keadaan kewangan anda sendiri sebelum melabur dalam mata wang kripto.'},
  {'role': 'user',
  'content': 'mengapa orang menggunakan blockchain untuk mengurus siasia digital',
  'content_ms': 'mengapa orang menggunakan blockchain untuk menguruskan sisa digital'},
  {'role': 'assistant',
  'content': 'There are several reasons why people use blockchain technology to manage digital assets:\n\n1. Security: Blockchain technology provides a high level of security for managing digital assets. The decentralized nature of blockchain makes it difficult for hackers to manipulate or steal digital assets.\n2. Transparency: Blockchain provides a transparent record of all transactions related to a digital asset. This makes it easier to track ownership and verify the authenticity of the asset.\n3. Accessibility: Blockchain technology allows for the easy transfer and management of digital assets, making them accessible to anyone with an internet connection.\n4. Immutability: Once data is recorded on a blockchain, it cannot be altered or deleted. This ensures that the record of ownership and transactions related to a digital asset is permanent and cannot be tampered with.\n5. Programmability: Blockchain technology allows for the creation of smart contracts, which are self-executing contracts with the terms of the agreement directly written into code. This enables automatic and secure transfer of digital assets based on predefined conditions.\n6. Interoperability: Blockchain technology enables the interoperability of different digital assets and systems, allowing for seamless transfer and exchange of value.\n\nOverall, the use of blockchain technology for managing digital assets provides a secure, transparent, and decentralized system that is accessible to anyone with an internet connection. This makes it an attractive solution for managing various types of digital assets, including cryptocurrencies, digital art, and other digital assets.',
  'content_ms': 'Terdapat beberapa sebab mengapa orang menggunakan teknologi blockchain untuk mengurus aset digital:\n\n1. Keselamatan: Teknologi Blockchain menyediakan tahap keselamatan yang tinggi untuk mengurus aset digital. Sifat blockchain yang tidak berpusat menyukarkan penggodam untuk memanipulasi atau mencuri aset digital.\n2. Ketelusan: Blockchain menyediakan rekod yang telus bagi semua transaksi yang berkaitan dengan aset digital. Ini memudahkan untuk menjejak pemilikan dan mengesahkan ketulenan aset.\n3. Kebolehcapaian: Teknologi Blockchain membolehkan pemindahan dan pengurusan aset digital dengan mudah, menjadikannya boleh diakses oleh sesiapa sahaja yang mempunyai sambungan internet.\n4. Ketidakbolehubahan: Setelah data direkodkan pada rantaian blok, ia tidak boleh diubah atau dipadamkan. Ini memastikan rekod pemilikan dan transaksi yang berkaitan dengan aset digital adalah kekal dan tidak boleh diusik.\n5. Kebolehprograman: Teknologi Blockchain membolehkan penciptaan kontrak pintar, yang merupakan kontrak yang dilaksanakan sendiri dengan syarat perjanjian yang ditulis terus ke dalam kod. Ini membolehkan pemindahan aset digital secara automatik dan selamat berdasarkan syarat yang telah ditetapkan.\n6. Saling kendali: Teknologi Blockchain membolehkan kesalingoperasian aset dan sistem digital yang berbeza, membolehkan pemindahan dan pertukaran nilai yang lancar.\n\nSecara keseluruhannya, penggunaan teknologi blockchain untuk mengurus aset digital menyediakan sistem yang selamat, telus dan terpencar yang boleh diakses oleh sesiapa sahaja yang mempunyai sambungan internet. Ini menjadikannya penyelesaian yang menarik untuk mengurus pelbagai jenis aset digital, termasuk mata wang kripto, seni digital dan aset digital lain.'}],
'filename': 'output-audio/1-0-0.mp3'}
\end{lstlisting}

The audio synthetic dataset and implementation published at \href{https://huggingface.co/datasets/mesolitica/malaysian-youtube-audio-instructions}{mesolitica/malaysian-youtube-audio-instructions}.

\section{Synthetic Visual Malaysian Context Dataset}

To enhance the dataset for visual question answering (QA) tasks, we embarked on a comprehensive curation process aimed at capturing the essence of Malaysian daily life and surroundings. To achieve this, we meticulously gathered a vast array of local context images that encapsulate the cultural, social, and environmental facets unique to Malaysia. Our dataset spans a broad spectrum of themes, encompassing everything from the tantalizing diversity of Malaysian cuisine to the intricate details of local transportation systems and road infrastructure. Leveraging reputable sources such as resepichenom.com, wapcar.my, and motomalaysia.com, we meticulously collected images depicting iconic Malaysian foods, cars, motorcycles, road signs, and driving scenarios.

Each category was carefully selected to provide a holistic representation of Malaysian life, ensuring that our dataset captures the nuances and intricacies inherent to the local culture. For instance, the images of delectable Malaysian dishes sourced from resepichenom.com offer a glimpse into the rich culinary heritage of the nation, featuring beloved classics and regional specialties alike. Similarly, the inclusion of images showcasing Malaysian vehicles and road signs from wapcar.my, motomalaysia.com, and MS Wikipedia adds another layer of authenticity, allowing the model to familiarize itself with the unique characteristics of local transportation and traffic regulations.

Furthermore, our dataset extends beyond static imagery to incorporate dynamic elements, such as online driving tests, providing real-world scenarios and challenges that reflect the everyday experiences of Malaysian motorists. By compiling this diverse and extensive collection of visual content, we aim to provide our model with a rich and immersive learning experience, enabling it to understand and interpret Malaysian contexts with greater accuracy and depth. We designed a synthetic question-answering dialogues between assistant and a person about the photo. The answers are based on captions associated with the original data of the image.

An example of generated dataset are as follow,

\begin{lstlisting}[breaklines=true]
{
  "image": "pic/Yamaha NMAX (2016)_1.jpg",
  "image_url": "https://www.motomalaysia.com/wp-content/uploads/2017/10/yamaha-nmax-2.jpg",
  "conversations": [
    {
      "role": "user",
      "content": "<image>Ini model apa?"
    },
    {
      "role": "assistant",
      "content": "Yamaha NMAX (2016)"
    },
    {
      "role": "user",
      "content": "apa spesifikasi moto tu?"
    },
    {
      "role": "assistant",
      "content": "Manufacturer: Yamaha\nModel: 2016\nGeneration: Yamaha NMAX (2016)\n\nType: Scooter\nColour Available: Grey, Red\n\nDimension: 1,955 x 740 x 1,115 mm\nWheelbase: 1,350 mm\nWeight: 136 kg\nSeat Height: 765 mm\nGround Clearance: 135 mm\nMeter Display: Digital - Speedometer, Fuel Gauge\n\nEngine: 4-Stroke, SOHC\nDisplacement: 155 CC\nNo Of Cylinders: 1-Cylinder\nCompression Ratio: 10.5 : 1\nBore x Stroke: 58mm x 58.7mm\nMax Power: 11.1 kW @ 8,000 RPM\nMax Torque: 14.4 Nm @ 6,000 RPM\nTransmission: Automatic, CVT\nClutch: Dry, centrifugal, Shoe\nIgnition System: Transistor Coil Ignition (TCI)\nStarting System: Electric Starter\nCooling System: Liquid Cooled\nFuel System: Fuel Injection\nFuel Capacity: 6.6 Liters\n\nFeatures:\nType: Underbone\nFront Suspension: Telescopic\nRear Suspension: Unit Swing\nFront Brake: Disc\nRear Brake: Disc\nFront Tire: 110/70-13\nRear Tire: 130/70-13"
    }
  ]
}
\end{lstlisting}

The synthetic visual dataset and its implementation can be found on the HuggingFace Collections page under \href{https://huggingface.co/collections/mesolitica/multimodal-malaysian-llm-dataset-653a16214037a1bc4417eb3a}{Multimodal Malaysian LLM dataset}.

\section{Synthetic Data Generation for Multi-Images, Multi-Audio Multi-turn Instructions}

We address a notable gap observed in existing open-sourced multimodal datasets, as they lack the capability to replicate the multi-image or multi-audio interactions achieved by OpenAI's ChatGPT4. To address this limitation, we undertook the creation of a dataset that encompasses multiple images and audio segments related within a single multiturn conversation context.

Our methodology began with the assembly of pairs of images, audios or a mix of audios and image each accompanied by its respective description. We leveraged Mistral to generate the dataset for relationship data. The model was then tasked with determining potential relationships between these pairs.

By adopting this approach, we effectively simulated scenarios involving multi-image and multi-audio contexts in multiturn conversations. This simulation serves to enhance the capacity of multimodal models to learn from and respond to diverse media inputs, thereby advancing their capabilities beyond single-media-focused learning paradigms and a single turn conversation.

\subsection{Synthetic Multi-Images Relationship Instruction}

To facilitate the generation of synthetic multi-images visual question answering (QA) data, we employed a systematic approach leveraging existing image captions from the pretrained LLAVA dataset \cite{liu2023visual}. Our methodology involved creating prompts by posing queries related to the relationship between two different image captions. Subsequently, we utilized the Mistral-7B-Instruct-v0.2 model \cite{jiang2023mistral} for the generation of synthetic data, harnessing its capabilities to generate diverse and contextually relevant questions and answers based on the provided prompt below,

\begin{lstlisting}[breaklines=true]
  pic1 = 'this is the caption for the first picture'
  pic2 = 'this is the caption for the second picture'
  prompt = f"""
  Picture 1: {pic1} 
  Picture 2: {pic2}
  What is related between picture 1 and picture 2.
  """
  mistral(prompt)
\end{lstlisting}

An example of generated dataset are as follow,

\begin{lstlisting}[breaklines=true]
  {'filename': ['00370/003705168.jpg', '00611/006111738.jpg'],
 'filename_description': ['paco man edp gift set',
  "raspberry lemonade is a must in new york's famous cocktail scene"],
 'instruction': 'What is related between picture 1 and picture 2',
 'answer': "There is no direct relation between Picture 1 (Paco Rabanne 1 Million Edp Gift Set) and Picture 2 (Raspberry Lemonade being a must in New York's famous cocktail scene). Both are unrelated to each other. The first picture is an image of a perfume gift set, while the second picture is a depiction of a popular cocktail in New York City.",
 'instruction_ms': 'Apakah yang berkaitan antara gambar 1 dan gambar 2',
 'answer_ms': 'Tiada hubungan langsung antara Gambar 1 (Paco Rabanne 1 Million Edp Gift Set) dan Gambar 2 (Raspberry Lemonade menjadi must dalam adegan koktel terkenal di New York). Kedua-duanya tidak berkaitan antara satu sama lain. Gambar pertama ialah imej set hadiah minyak wangi, manakala gambar kedua ialah gambaran koktel popular di New York City.'}
\end{lstlisting}

All synthetic dataset and implementation published at \href{https://huggingface.co/datasets/mesolitica/synthetic-multiturn-multimodal#multi-images}{mesolitica/synthetic-multiturn-multimodal}.

\subsection{Synthetic Multi-Audio Relationship Instruction}

In our effort to produce synthetic multi-audio question answering (QA) datasets, we utilized pseudolabeled transcriptions sourced from ~\ref{sec:synthetic-audio}. These transcriptions served as the foundation for creating prompts that prompted queries regarding the relationship between two distinct audio transcriptions. Leveraging the capabilities of the Mistral-7B-Instruct-v0.2 model \cite{jiang2023mistral}, we effectively generated synthetic data by formulating diverse and contextually relevant questions and answers based on the prompt below,

\begin{lstlisting}[breaklines=true]
  audio1 = 'this is the transcription for audio 1'
  audio2 = 'this is the transcription for audio 2'
  prompt = f"""
  Audio 1: {audio1} 
  Audio 2: {audio2}
  What is related between audio 1 and audio 2
  """
  mistral(prompt)
\end{lstlisting}

An example of generated dataset are as follow,

\begin{lstlisting}[breaklines=true]
  {'filename': ['output-audio/3-2080-38.mp3', 'output-audio/0-2823-0.mp3'],
 'filename_description': ['Terima kasih Menteri. Saya jemput soalan tambahan yang kedua. Bagan Serai. Terima kasih Tuan Speaker. Berapakah jumlah kemalangan yang menyebabkan kematian disebabkan oleh pengaruh handphone, penggunaan handphone semasa mandu. Kerana guna handphone mandu ini dia macam mabuk lebih Tuan Speaker. Dan dia hilang orientasi. Dia tak tahu di mana traffic light, dia tak tahu dia di mana berada dan tiba-tiba dah sampai. Jadi apa kerajaan nak buat untuk menurunkan tabiat buruk menggunakan handphone semasa mandu.',
  'dalam video tu saya dah kitamkan kening lah sebab benda tu kita mencuba so at least kita dah mencuba kita kan nak mencuba kan masa ni lah mencuba kan janganlah pula usia macam aku dah 50 pun nak cuba kenapa masa buat lagu raya cover tu tak boleh hijau sebab dia nak image ketupat macam Aina Abdul juga dia ketupat kita bawa image rambut tu warna hijau ketupat juga kan tapi dah habis raya after this memang nak reveal jugalah kan habis ni memang saya akan kekalkan image yang very very formal je lah'],
 'instruction': 'What is related between audio 1 and audio 2',
 'answer': 'Audio 1 and Audio 2 are unrelated as they discuss different topics. In Audio 1, the speaker is discussing the issue of using handphones while driving and its contribution to accidents. In Audio 2, the speaker is talking about making a cover song for Raya and the challenges they faced in creating the image for the video.',
 'instruction_ms': 'Apakah yang berkaitan antara audio 1 dan audio 2',
 'answer_ms': 'Audio 1 dan Audio 2 tidak berkaitan kerana mereka membincangkan topik yang berbeza. Dalam Audio 1, penceramah membincangkan isu menggunakan fon tangan semasa memandu dan sumbangannya kepada kemalangan. Dalam Audio 2, penceramah bercakap tentang membuat lagu penutup untuk Raya dan cabaran yang mereka hadapi dalam mencipta imej untuk video itu.'}
\end{lstlisting}

This approach facilitated the development of comprehensive datasets comprising multi-audio QA pairs, enabling the training and evaluation of multimodal models capable of handling audio-based queries and responses.

All synthetic dataset and implementation published at \href{https://huggingface.co/datasets/mesolitica/synthetic-multiturn-multimodal#multi-audio}{mesolitica/synthetic-multiturn-multimodal}.

\subsection{Synthetic Image-Audio Relationship Instruction}

To create synthetic image-audio question answering (QA) datasets, we employed a combination of existing resources and data sources. Firstly, we utilized image captions obtained from the pretrained LLAVA dataset \cite{liu2023visual}, which provided a rich source of contextual information for images. Additionally, we leveraged pseudolabeled transcriptions sourced from the Malaya dataset, offering comprehensive representations of audio content. Using these assets, we formulated prompts that prompted inquiries regarding the relationship between specific image captions and corresponding audio transcriptions. Through the application of the Mistral-7B-Instruct-v0.2 model \cite{jiang2023mistral}, we generated synthetic data by crafting diverse and contextually relevant questions and answers based on the prompt below,

\begin{lstlisting}[breaklines=true]
  audio1 = 'this is the transcription for audio'
  picture1 = 'this is the caption for the image'
  prompt = f"""
  Audio 1: {audio1} 
  Picture 1: {image1}
  What is related between audio 1 and picture 1
  """
  mistral(prompt)
\end{lstlisting}

An example of generated dataset are as follow,

\begin{lstlisting}[breaklines=true]
  {'filename': ['output-audio/3-2648-47.mp3', '00180/001805101.jpg'],
  'filename_description': ['Saya mahu muka mereka terlihat beras, anda tahu apa yang saya maksudkan. Dan sanitizer. Dan kemudian ini adalah earphone. Sama-sama kalau airpod saya, anda tahu, hilang bateri. Saya tidak pasti jika saya patut membawa tripod saya kerana saya mungkin. Adakah saya akan melakukan TikTok di kafe? Saya tidak tahu tetapi tidak menyakiti untuk membawanya. Maksud saya, ia tidak begitu keras. Saya perlu membawa krim tangan saya. Dan kemudian bumbu. Dan lip balm. Dan kemudian kita siap untuk pergi.',
   'a water wheel with moss growing on the wheels metal print by randall white'],
  'instruction': 'What is related between audio 1 and picture 1',
  'answer': "The audio and picture do not have a direct relation to each other. The audio is about preparing items for an outing, including sanitizer, earphones, a tripod, and various other personal items. The picture is a print of a water wheel with moss growing on it by Randall White. There is no connection between the audio's content and the picture's subject matter.",
  'instruction_ms': 'Apakah yang berkaitan antara audio 1 dan gambar 1',
  'answer_ms': 'Audio dan gambar tidak mempunyai hubungan langsung antara satu sama lain. Audio adalah mengenai penyediaan item untuk keluar, termasuk pembersih, fon telinga, tripod dan pelbagai barangan peribadi lain. Gambar itu ialah cetakan roda air dengan lumut yang tumbuh di atasnya oleh Randall White. Tiada kaitan antara kandungan audio dan subjek gambar.'}
 
\end{lstlisting}

This approach enabled us to construct comprehensive datasets comprising image-audio QA pairs, facilitating the training and evaluation of multimodal models capable of handling queries spanning both image and audio modalities.

All synthetic dataset and implementation published at \href{https://huggingface.co/datasets/mesolitica/synthetic-multiturn-multimodal#multi-images-multi-audio}{mesolitica/synthetic-multiturn-multimodal}.

\subsection{Synthetic Multi-Images Multi-Audio in one session Instruction}

\begin{lstlisting}[breaklines=true]
paragraph = ' Audio 1: I hope she's on the same page with me because I feel like after years of being with her, she also is the same as me and it's money come to her account...    Image 1: new design inflatable game house with air blower.'
instruction = f'{paragraph}\n\nwhat is related between audio 1 and picture 1?'
mixtral(instruction)
\end{lstlisting}

An example of generated dataset,

\begin{lstlisting}[breaklines=true]
  {
    "filename": [
      "output-audio/3-2825-0.mp3",
      "00117/001170082.jpg"
    ],
    "filename_description": [
      "I hope she's on the same page with me because I feel like after years of being with her, she also is the same as me and it's money come to her account and then we agree that to put money inside because circumstances change. Like for example, her circumstance is better than me now because she works at Colgate. I don't know when I'm supposed to say that, but whatever, right? And she gets a good salary now and because of that good salary, she has no time and she's like, I don't know whether I should have taken this job. I don't even have time to like join my life already. But she earns more now. So like by right,",
      "new design inflatable game house with air blower"
    ],
    "instruction": "What is related between audio 1 and picture 1",
    "answer": "The relationship between the audio 1 and picture 1 is not clear without additional context. The audio is about a person's thoughts and feelings regarding their partner's financial situation and work commitments. The picture shows a new design inflatable game house with an air blower. There is no apparent connection between the two.",
    "instruction_ms": "Apakah yang berkaitan antara audio 1 dan gambar 1",
    "answer_ms": "Hubungan antara audio 1 dan gambar 1 tidak jelas tanpa konteks tambahan. Audio adalah mengenai pemikiran dan perasaan seseorang mengenai situasi kewangan pasangan mereka dan komitmen kerja. Gambar itu menunjukkan rumah permainan kembung reka bentuk baharu dengan pengtiup udara. Tiada hubungan yang jelas antara kedua-duanya."
  }
\end{lstlisting}

\subsection{Synthetic Multi-Images Multi-Audio in one session Instruction}

To generate multiple images, audio and dialogue turns that offers fine-grained interactions between multiple images, audio and words, mimicking user-assistant conversations.

We built a script that parse visual and audio context datasets and perform an algorithm that combine each visual context row into another audio context or image context row based on probability which makes our combined data more varied.

Below are the script to combine synthetic visual and audio context dataset,

\begin{lstlisting}[breaklines=true]
  llava_index = set(range(len(llava_en)))
  audio_index = set(range(len(audio_en)))
  combine = []
  llava = set()
  audio = set()
  for i in tqdm(range(N)):
    filename = []
    conversations = []
    for _ in range(random.randint(2, 4)):
        if random.random() > 0.4:
            img_id = random.choice(list(llava_index - llava))
            llava.add(img_id)
            s = llava_en[img_id]
        else:
            audio_id = random.choice(list(audio_index - audio))
            audio.add(audio_id)
            s = audio_en[audio_id]
        filename.extend(s['filename'])
        conversations.extend(s['conversations'])
    combine.append({
        'filename': filename,
        'conversations': conversations
    })
\end{lstlisting}

An example of combined synthetic visual and audio context datasets:

\begin{lstlisting}[breaklines=true]
  {'filename': ['/home/ubuntu/output-audio/0-3201-1.mp3',
  '/home/ubuntu/filtered-llava-images/000000416018.jpg',
  '/home/ubuntu/filtered-llava-images/000000128752.jpg',
  '/home/ubuntu/filtered-llava-images/000000501015.jpg'],
 'conversations': [{'role': 'user',
   'content': "<audio>What is Tasha's opinion on the spiciness of the first soup served?"},
  {'role': 'assistant',
   'content': 'Based on the conversation, Tasha felt that the first soup served was too spicy.'},
  {'role': 'user', 'content': 'Okay. Mengapa Tasha pandangan begini?'},
  {'role': 'assistant',
   'content': 'The exact reason for Tasha\'s opinion is not mentioned in the conversation, but she can be heard saying "Yup. Yes. Ni pedas gila." which suggests that the first soup served was extremely spicy.'},
  {'role': 'user',
   'content': 'Okay. Sekarang kita tanya Jane. Okay. Jane, how spicy is your soup?'},
  {'role': 'assistant',
   'content': 'Based on the conversation, Jane said "Second pedas" which means that she found the spiciness of her soup to be less spicy compared to the first soup.'},
  {'role': 'user',
   'content': "What implications does the presence of the steering wheel controller have for the type of game they're playing?\n<image>"},
  {'role': 'assistant',
   'content': 'Given that one of the people in the image is holding a video game controller shaped like a steering wheel, it is highly likely that they are playing a racing or driving simulation game. The steering wheel controller is specifically designed to provide a more immersive and realistic gaming experience for games that involve controlling vehicles. This suggests that the game they are engaging with is focused on car or vehicle navigation, and the participants are actively controlling in-game actions by manipulating the steering wheel controller.'},
  {'role': 'user',
   'content': 'What factors could affect the flight of this aircraft?\n<image>'},
  {'role': 'assistant',
   'content': "Several factors could affect the flight of the British Airways jetliner observed in the image. These may include:\n\n1. Weather conditions: The plane is flying through a foggy sky, which may impact visibility, turbulence, and overall flight conditions. Adverse weather events, such as storms or strong winds, could also significantly impact the flight.\n\n2. Air traffic: The presence of other aircraft in the airspace could influence the flight route, altitude, or speed. Air traffic controllers need to manage multiple aircraft in the sky to maintain safety and avoid conflicts.\n\n3. Technical issues: Aircraft systems and equipment must function properly to ensure a safe and efficient flight. Any technical issues or malfunctions may disrupt the flight and require prompt action from the crew.\n\n4. Human factors: The skills, experience, and decision-making processes of pilots and other crew members can significantly impact the safety, comfort, and efficiency of the flight.\n\n5. Regulatory constraints: International aviation regulations and guidelines from governing bodies can affect the aircraft's flight trajectory, speed, altitude, and other operational aspects.\n\nConsidering these factors, the British Airways jetliner needs to navigate through the foggy sky efficiently while maintaining safety protocols and adhering to aviation regulations."},
  {'role': 'user',
   'content': '<image>\nHow would you describe the positioning of the bottles in this image?'},
  {'role': 'assistant',
   'content': "In the image, there are several bottles positioned in various locations. Some bottles are inside the refrigerator, with one being held in the person's hand, while others are placed outside of the fridge. The person is posing near the open refrigerator, which appears to have quite a few bottles arranged in it. The bottles inside the fridge are lined up horizontally on different shelves at various heights, indicating that they might be chilling until they are ready to be consumed. There are also a couple of bottles placed outside the fridge, possibly on a countertop or other surfaces within the image. The dining table and a laptop can also be seen in the background, but they are not directly related to the positioning of the bottles."}]}
\end{lstlisting}

By doing this approach, multimodal model would enhance its capabilities on understanding and reasoning across multiple images, audios and dialogue turns.

The implementation can be found on the Github repository page at \href{https://github.com/mesolitica/multimodal-LLM/tree/master/prepare-dataset}{mesolitica/multimodal-LLM/prepare-dataset}.

\section{Finetuning Procedure}

MMMModal aims to align visual and audio information from pretrained vision and audio encoders with an advanced large language model (LLM). We aim to bridge the gap between the visual and audio encoders and the LLM using a linear projection layer. To create an effective multimodal model, we followed a two-stage training approach exemplified by the works of \cite{han2023chartllama,zhu2023minigpt4,liu2024hidden} which have notably produced great results. In the initial stage, the model is pretrained on aligned image-text pairs and audio-text pairs to acquire knowledge of vision and audio language through the alignment projection layer. In the second stage, we fine-tune the pretrained model using a generated multiturn multiaudio images synthetic dataset, incorporating a designed conversational template to enable model comprehension on multi-images, multi-audio, and multi-images-multi-audio within a single multiturn session.

The visualization below provides an overview of the architecture for MMMModal,

\begin{figure}[hbt!]
  \centering
  \includegraphics[width=0.6\linewidth]{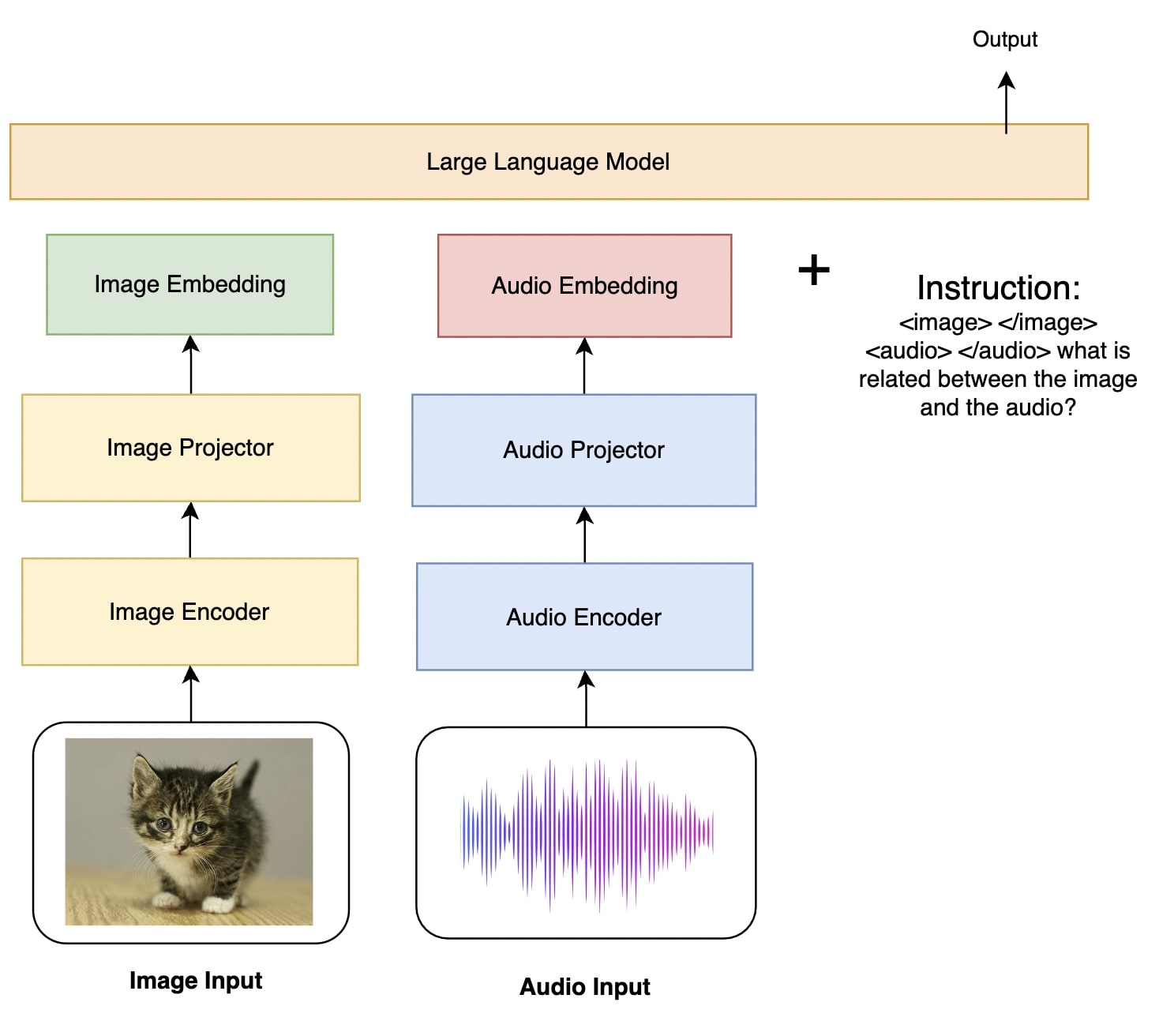} % Replace with your image file
\end{figure}

\subsection{Pretraining for Visual Feature Alignment}\label{sec:visual-alignment}

During the initial pretraining stage, the primary objective is to equip the model with the ability to understand and generate language based on visual input. This is achieved through exposure to a diverse set of paired image-text data, where the model learns to associate visual information with corresponding textual descriptions. In our work, we utilize the pre-trained SigLIP encoder \cite{zhai2023sigmoid} to extract visual features for input into the projection layer. This projection layer facilitates the connection of image features into the text embedding space. The output of this projection layer then acts as the input to the Large Language Model, instructing it on how to generate appropriate textual responses based on the visual features provided. Only the linear projection layer is pretrained during the whole pretraining procedure; the pretrained vision encoder and the LLM stay frozen.

In our approach, we adopt the same projection layer as LLAVA \cite{liu2023visual}, which consists of two hidden layers with GELU activation at the middle. However, we introduce two new tokens, \texttt{<image>} and \texttt{</image>}, to facilitate the incorporation of visual information. These tokens serve as markers to indicate the beginning and end of projected visual output, enabling seamless integration within the text embedding.

\newpage

The visualization below illustrates the process of inserting projected visual output between the \texttt{<image>} and \texttt{</image>} tokens, enhancing the model's ability to handle visual inputs effectively.

\begin{figure}[hbt!]
  \centering
  \includegraphics[width=0.6\linewidth]{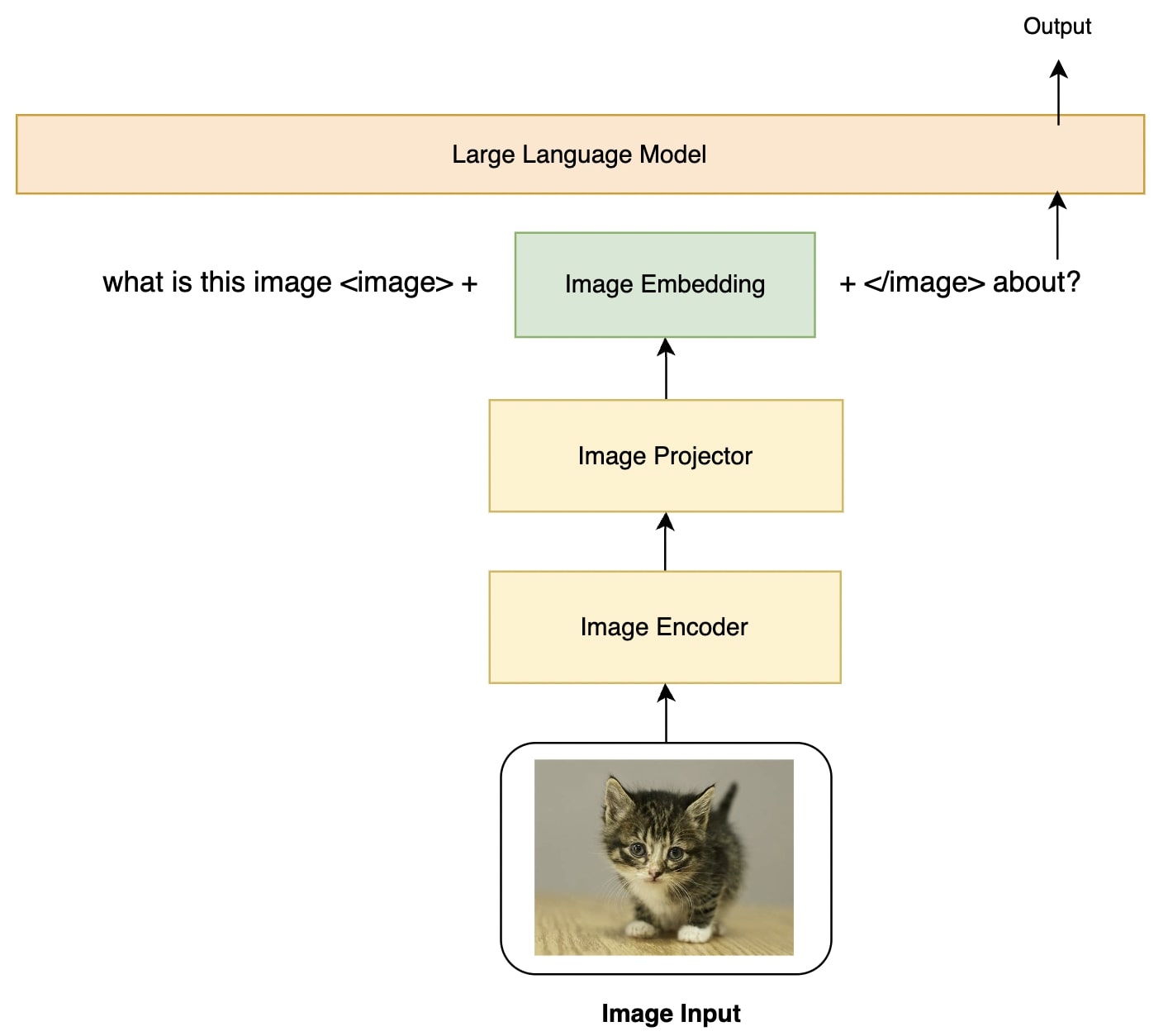} % Replace with your image file
\end{figure}

The implementation for visual feature alignment pretraining can be found \href{https://github.com/mesolitica/multimodal-LLM/tree/master/vision-only}{here}.

\subsection{Pretraining for Audio Feature Alignment}\label{sec:audio-alignment}

We also want to equip the model with the capability to comprehend and produce language from audio input. This is achieved by exposing the model to a diverse set of paired audio-text datasets, allowing it to learn the correspondence between audio features and corresponding textual descriptions. We utilize the pre-trained Whisper encoder to extract audio features for input into the projection layer. The injected projection layer plays a pivotal role in this process, serving as a bridge between the audio and text domains. The output of this projection layer serves as input to the Large Language Model, guiding it in generating appropriate textual responses based on the audio features provided. It is important to note that while the linear projection layer is trained throughout the entire pretraining procedure, the pretrained audio encoder and the Large Language Model remain static, or "frozen." This ensures that the model focuses specifically on learning the associations between auditory features and textual information without altering the underlying representations learned in the audio encoder or the language model.

In our method, we draw inspiration from LLAVA's \cite{liu2023visual} projection layer but introduce modifications to handle longer sequences more efficiently. Instead of using two linear layers, we opt for a convolutional layer with a kernel size of 40 and a stride size of 3 for the first layer. The output is then passed through a linear layer with a GELU activation function at its midpoint. This approach helps us reduce the sequence length by a factor of 3, which is crucial given that the Whisper encoder outputs sequences of length 1500. Additionally, we introduce two new tokens, \texttt{<audio>} and \texttt{</audio>}, serving as markers for projected audio output, the same procedure as ~\ref{sec:visual-alignment}.

\newpage

The visualization below demonstrates the procedure of embedding projected audio output between the \texttt{<audio>} and \texttt{</audio>} tokens.

\begin{figure}[hbt!]
  \centering
  \includegraphics[width=0.6\linewidth]{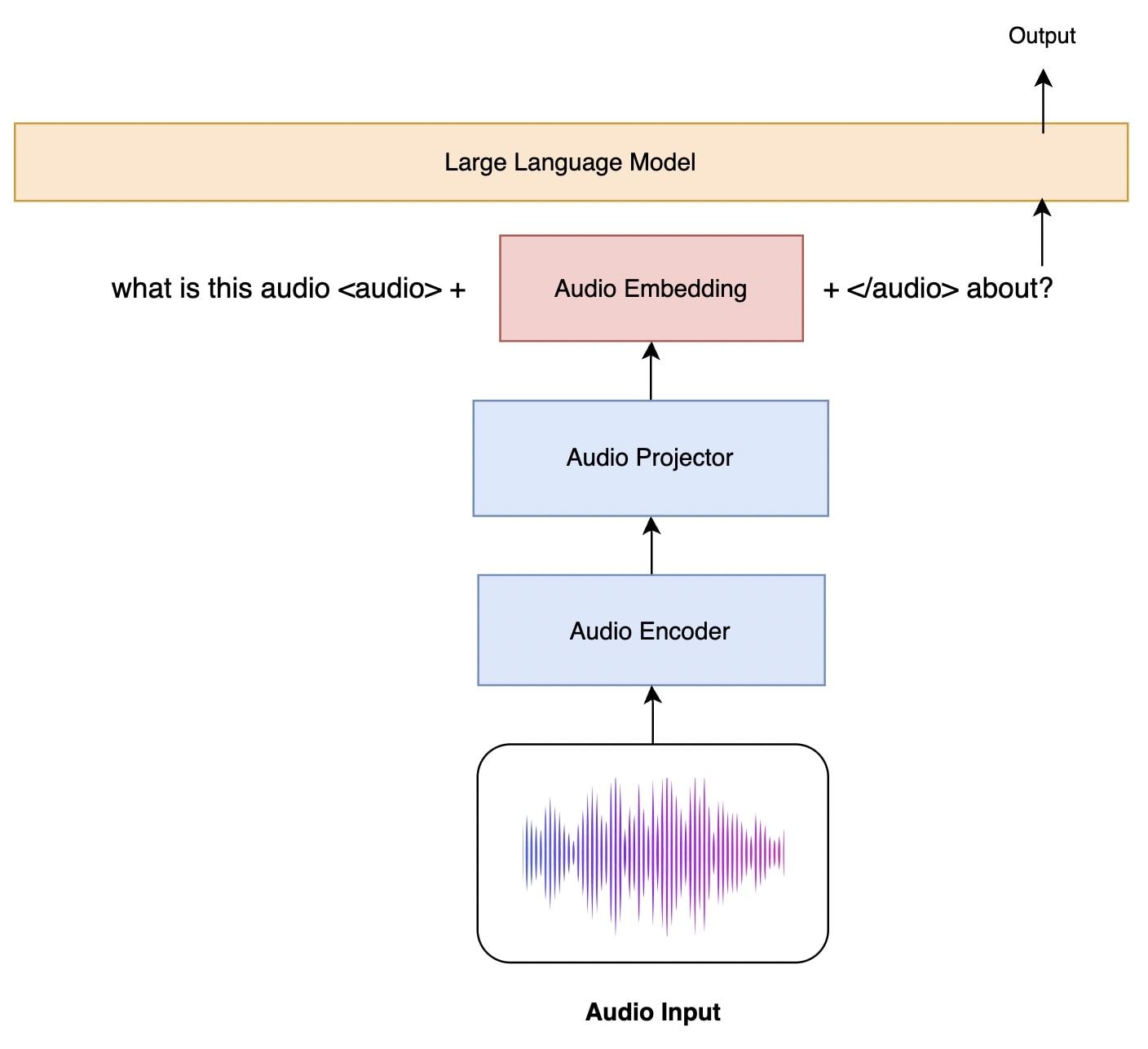} % Replace with your image file
\end{figure}

The implementation for audio feature alignment pretraining can be found \href{https://github.com/mesolitica/multimodal-LLM/tree/master/audio-only}{here}.

\subsection{Instruction Finetuning}

Following the initial pre-training alignment phase, the projection layer will be able to generate image and audio features that are effectively aligned with the pretrained Large Language Model text embedding space. However the pretrained LLM may still struggle to provide effective responses to human inquiries and may be unable to comprehend instructions involving multiple images, multiple audio files, and combinations thereof. To address this, we implement a second stage, utilizing the generated synthetic data to enhance performance and refine the alignment between audio, visual embeddings, the LLM, and instructions.

This stage involves fine-tuning the projection layer and the LLM on our collection corpus of generated synthetic data encompassing multiturn, multi-images and multi-audios data. The objective is to enable our multimodal model to process multiple inputs from images or audios and engage in multiturn conversation seamlessly by implementing an efficient span fill-in mechanism anchored around the \texttt{<image>} and \texttt{<audio>} tokens, the same procedure as ~\ref{sec:visual-alignment} and ~\ref{sec:audio-alignment}.

The hyperparameters involves in this finetuning stage are detailed below:

\begin{table}[h]
  \centering
  \begin{tabular}{lccl}
    \hline
    \textbf{Hyperparameter} & \textbf{Value} \\
    \hline
    DeepSpeed               & ZeRO-2 Offload \\
    Batch Size              & 12             \\
    Learning Rate           & constant 2e-5  \\
    Precision               & bfloat16       \\
    \hline
  \end{tabular}
\end{table}

Complete fine-tuning 8192 context length implementation at \href{https://github.com/mesolitica/multimodal-LLM/blob/master/run-deepspeed.sh}{here}.

\section{Training Discussion}

\subsection{Pretraining Alignment}

In our exploration of visual alignment pretraining, we conducted experiments using both \href{https://huggingface.co/google/siglip-base-patch16-384}{google/siglip-base-patch16-384} and \href{https://huggingface.co/google/siglip-large-patch16-384}{google/siglip-large-patch16-384} architectures. Surprisingly, we observed no discernible difference in terms of loss between the two variants. Similarly, in another set of experiments, we compared the use of SigLIP \cite{zhai2023sigmoid} output with SigLIP \cite{zhai2023sigmoid} hidden layer output for visual alignment pretraining. Intriguingly, while both approaches yielded similar loss outcomes, discrepancies emerged during instruction finetuning. Specifically, utilizing SigLIP \cite{zhai2023sigmoid} output resulted in a significantly higher incidence of hallucinated responses. However, it's essential to note that these findings lack empirical backing, necessitating further investigation to validate their significance.

In contrast, our exploration of audio alignment did not entail a comparative study, primarily due to resource constraints. Instead, we relied on the output from the Malaysian Whisper Small model's hidden layers, informed by insights gleaned from the SigLIP encoder \cite{zhai2023sigmoid}.

\subsection{Instruction Finetuning}

During distributed multi-GPU training, data shuffling may inadvertently result in certain devices receiving instruction samples containing only audio or images, leading to discrepancies in gradient computation across devices. This inconsistency can manifest as "parameters that were not used in producing loss," particularly problematic when employing frameworks like Torch Elastic or DeepSpeed Zero 2, where such discrepancies can cause training to stall indefinitely. To mitigate this issue, we have implemented a solution wherein empty image and audio placeholders are appended to each sample, ensuring that every device receives both modalities. Additionally, attention masks are set to zero for these placeholders, effectively disregarding them during computation. This approach ensures uniformity in gradient calculation across devices, preventing training interruptions and improving overall model stability during distributed training sessions.

Importantly, during inference, we can safely omit the addition of these empty placeholders as we have already configured the model to ignore them by setting the corresponding labels appropriately.

\section{Examples}

This section presents examples that highlight the model's capacity to comprehend and produce responses relating to visual and audio input, showcasing the efficacy and potential of our proposed MMModal. These examples clearly demonstrate how the model handles and combines various information modalities, including audio and pictures.

\begin{table}[hbt!]
  \centering
  \setlength{\extrarowheight}{3pt} % Adjust the value as needed
  \renewcommand{\arraystretch}{1.5} % Adjust the value as needed
  \begin{tabular}{p{6.5cm}p{6.5cm}}
    \hline
    \textbf{Multi Images Input Example}                                                                                                                                                                                                                                                      \\[6pt]
    \hline
    \hline
    \\[-1pt]
    \includegraphics[width=0.9\linewidth,height=4cm]{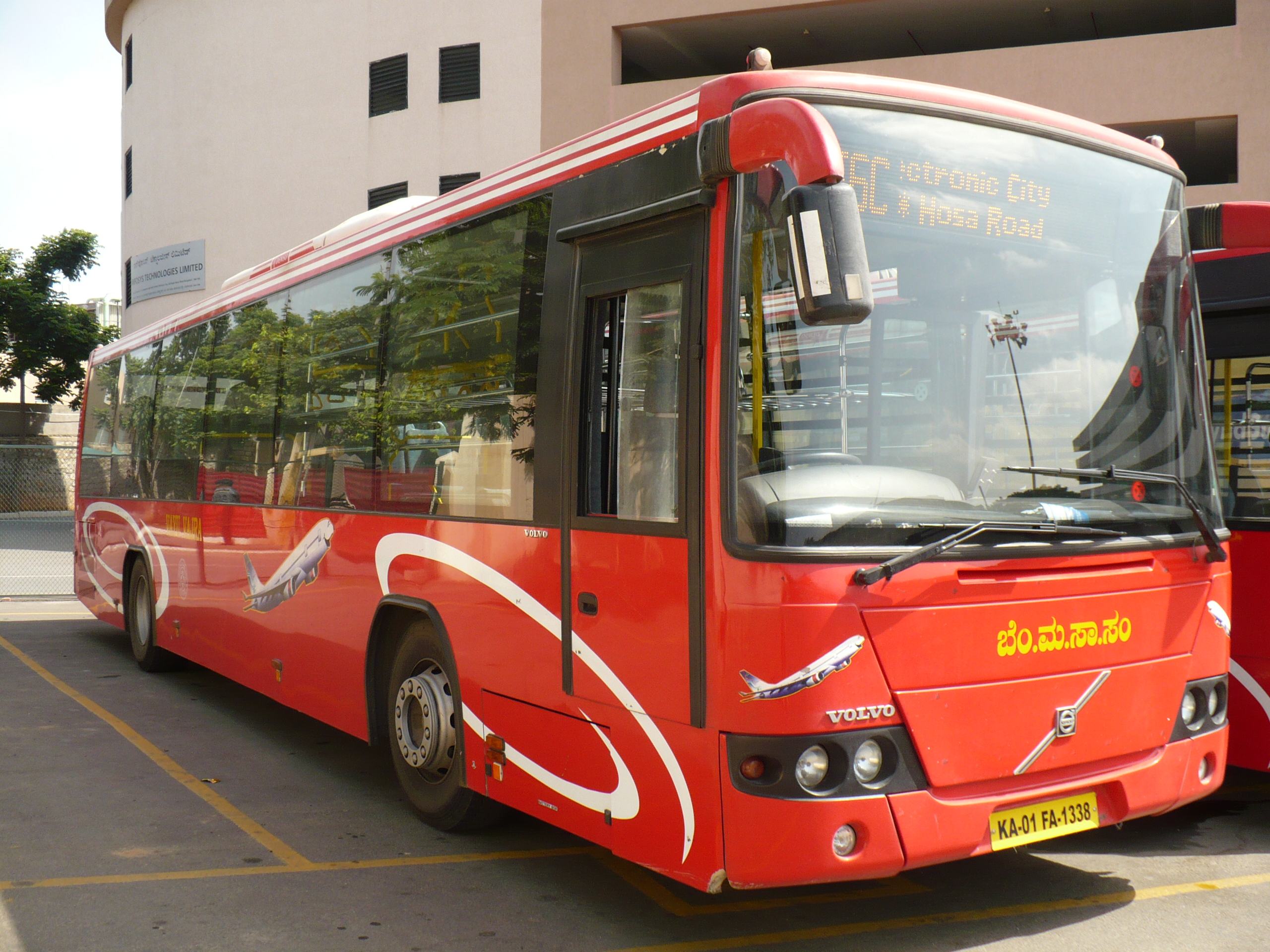} & \includegraphics[width=0.9\linewidth,height=4cm]{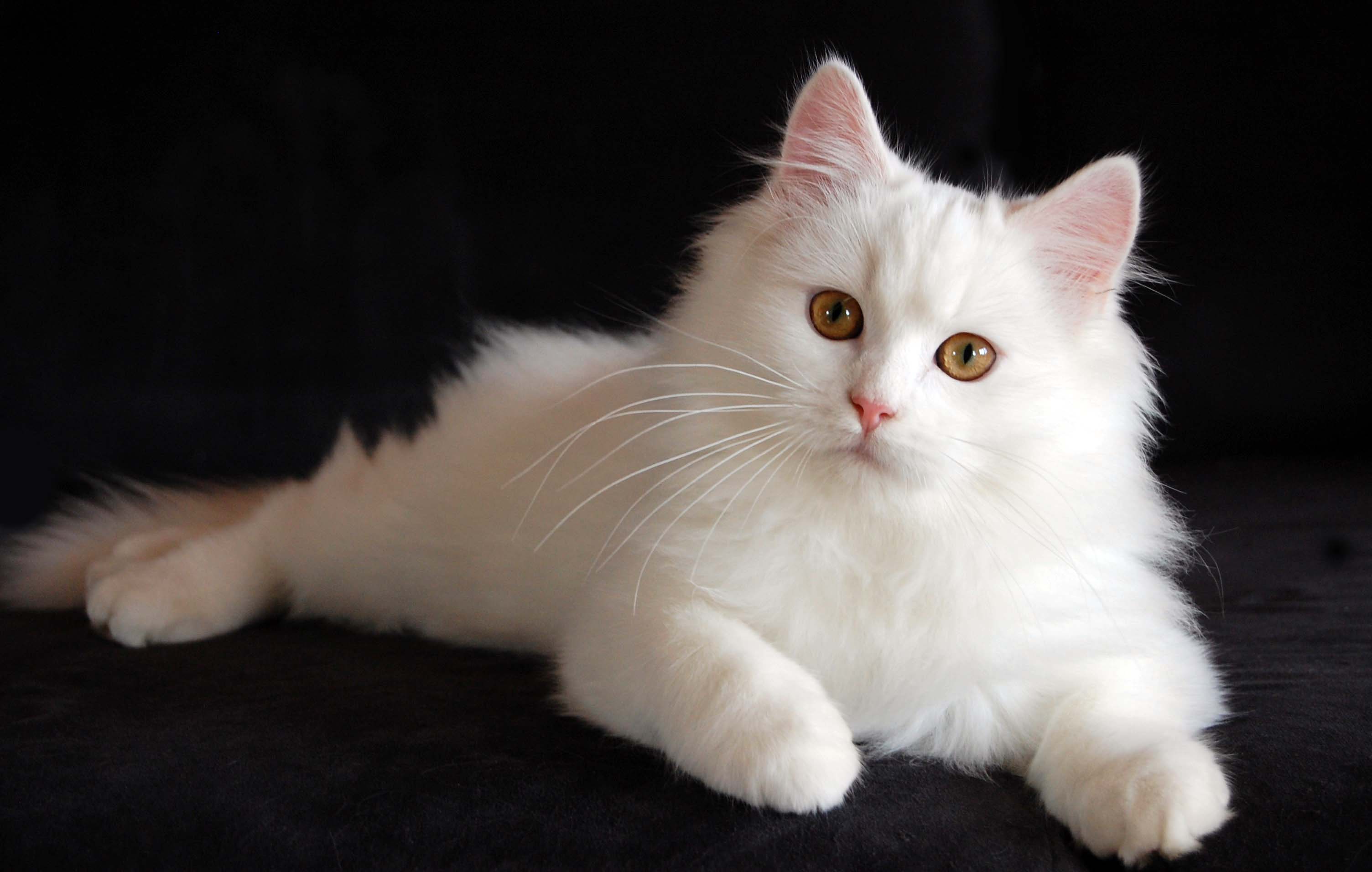}                                                                                                                                                           \\
    \\[-1pt]
    \hline
    User                                                           & <image> </image> <image> </image> What is related between those images?                                                                                                                                                 \\
    MMMModal                                                       & There is no direct relationship between image 1 and image 2. The first image shows a tour bus driving down a street, while the second image shows a white cat lying on a black couch. Both are unrelated to each other. \\
    \hline
  \end{tabular}
\end{table}

\begin{table}[hbt!]
  \setlength{\extrarowheight}{3pt}
  \renewcommand{\arraystretch}{1.5}
  \begin{tabular}{p{6.5cm} p{6.5cm}}
    \hline
    \textbf{Multi Audio Input Example}                                                                                                                                                                                                                                                                                                                                                                                                                                                                                                                                                                                                                                                                                                                                                                                                                                                                                                                                                                                                                                                                                                                                                                                                                                                                                                                                                                                                                                                                                                                                                                                                                                                                            \\[3pt]
    \hline
    \hline
    Audio 1 Transcription: anda tahu keuntungan boleh lebih tinggi daripada keuntungan kewangan rumah maka saya tidak akan mencari dalam akaun saya akan mencari ke dalam ethereum atau beberapa crypto punks bergantung pada faktor risiko anda kerana rumah kajang dihantar tidak mengganggu dsr saya sejauh ini jadi sekarang apa posisi saya untuk mendapatkan kewangan ketiga jadi mungkin setelah melihat sekeliling saya menemui seorang penjual yang dapat menutupi perhubungan tetapi bank hanya menerima 70\% dari itu saya boleh membayar perbezaan dengan menggunakan wang ini kerana sekali lagi ia menyusahkan saya dan aset tetapi jika anda tidak selesa dengan mencari. & Audio 2 Transcription: Sekarang Adria nak tunjukkan dia punya kotak Kotak dia simple, dia transparent Vib Max dekat sini, root beer Dia punya flavour punya nama Dekat atas dia tulis root beer float Bagi korang senang sikit nak tengok Tepi sini 12,000 puff Lepas tu dekat sebelah sini dia ada tulis lah dekat sini 22ml untuk 12,000 puff lah Rechargeable type C, 600mAh Adjustable airflow So dekat atas ada satu seal Korang just bukakan Untuk dia punya silicon actually agak besar Masa korang ambil keluar device yang baru kan Dia ada 2 sticker, 1 silicon                                                                                                                                                                                                                                                                                                                                                                                                                                              \\
    \hline
    User                                                                                                                                                                                                                                                                                                                                                                                                                                                                                                                                                                                                                                                                                 & <audio> </audio> <audio> </audio> What is related for both audio?                                                                                                                                                                                                                                                                                                                                                                                                                                                                                                                                                                                                                                                                                                                                                                                                                                                                                                                                                      \\
    MMMModal                                                                                                                                                                                                                                                                                                                                                                                                                                                                                                                                                                                                                                                                             & Both audio clips are discussing the topic of buying a product, possibly a vaping device, and providing information about its features and specifications. In Audio 1, the speaker is discussing the decision to invest in cryptocurrencies instead of a house loan due to the potential higher returns. In Audio 2, the speaker is describing the features of a vaping device, including its transparency, simple design, and adjustable airflow. The speaker also mentions the device's charging capabilities and the amount of puffs it can provide. The connection between the two audio clips is that they both discuss the importance of making informed decisions when purchasing a product. In Audio 1, the speaker is discussing the decision to invest in cryptocurrencies, while in Audio 2, the speaker is discussing the features and specifications of a vaping device. Both clips emphasize the importance of understanding the product's features and specifications before making a purchase decision. \\
    \hline
  \end{tabular}
\end{table}

\begin{table}[hbt!]
  \centering
  \setlength{\extrarowheight}{3pt}
  \renewcommand{\arraystretch}{1.5}
  \begin{tabular}{>{\arraybackslash}m{6.5cm}>{\arraybackslash}m{6.5cm}}
    \hline
    \textbf{Images Audio Input Example}                                                                                                                                                                                                                                                                                                                                                                                                                                                                                                                                                                                                                                                                                                       \\[6pt]
    \hline
    \hline
    \includegraphics[width=0.9\linewidth]{pic/cat.jpg} & Audio 1 Transcription:  anda tahu keuntungan boleh lebih tinggi daripada keuntungan kewangan rumah maka saya tidak akan mencari dalam akaun saya akan mencari ke dalam ethereum atau beberapa crypto punks bergantung pada faktor risiko anda kerana rumah kajang dihantar tidak mengganggu dsr saya sejauh ini jadi sekarang apa posisi saya untuk mendapatkan kewangan ketiga jadi mungkin setelah melihat sekeliling saya menemui seorang penjual yang dapat menutupi perhubungan tetapi bank hanya menerima 70\% dari itu saya boleh membayar perbezaan dengan menggunakan wang ini kerana sekali lagi ia menyusahkan saya dan aset tetapi jika anda tidak selesa dengan mencari \\
    \hline
    User                                               & What is related between the image and audio?                                                                                                                                                                                                                                                                                                                                                                                                                                                                                                                                                                                                                                         \\
    MMMModal                                           & There is no direct relationship between image 1 and audio 1. The image shows a white cat lying on a black couch, while the audio is about a person discussing their decision to invest their money in cryptocurrencies instead of a housing loan.                                                                                                                                                                                                                                                                                                                                                                                                                                    \\
    \hline
  \end{tabular}
\end{table}

\clearpage

\section{Future Work}

In our future work, we aim to enhance our capabilities by focusing on several key areas. Firstly, we intend to refine our approach to generating synthetic datasets that incorporate multi-images and multi-audio inputs. This will involve expanding the dataset to include more complex relationships between inputs and facilitating comparisons involving more than two inputs. Additionally, we recognize the importance of incorporating a wider range of visual Malaysian context datasets into our model training pipeline. By diversifying our data sources, we can ensure that our model is equipped to handle a broader array of real-world scenarios and contexts, ultimately improving its performance and relevance in practical applications.

We also aim to extend our multimodal model to incorporate video inputs. This involves leveraging any video encoder and pretraining video alignment to enhance the model's ability to comprehend video data. By incorporating video understanding capabilities, our model can offer richer insights and responses across various multimedia contexts.

\section{Acknowledgement}

Special thanks to Malaysia-AI volunteers especially \href{https://www.linkedin.com/in/wan-adzhar-faiq-adzlan-19a27baa/}{Wan Adzhar Faiq Adzlan}, \href{https://www.linkedin.com/in/ammar-azman/}{Ammar Azman}, \href{https://www.linkedin.com/in/amzar96/}{M. Amzar}, \href{https://www.linkedin.com/in/muhammad-farhan-helmy-0529501a7/}{Muhammad Farhan}, \href{https://www.linkedin.com/in/syafie-nizam/}{Syafie Nizam}, \href{https://www.linkedin.com/in/halimshukor/}{Halim Shukor}, \href{https://www.linkedin.com/in/alif-aiman-1b334b24b/}{Alif Aiman}, \href{https://www.linkedin.com/in/azwan-zuharimi/}{Azwan Zuharimi} and \href{https://www.linkedin.com/in/haziqzikry/}{Haziq Zikry} for contributing dataset to train MMMModal.

We would like to express our gratitude to NVIDIA Inception for generously providing us with the opportunity to train our model on the Azure cloud. Their support has played a crucial role in the success of our research, enabling us to leverage advanced technologies and computational resources.

We extend our thanks to the wider research community for their valuable insights and collaborative discussions, which have greatly influenced our work. This paper reflects the collective efforts and contributions from both NVIDIA Inception and the broader research community.

\section{Conclusion}

In this paper, we introduce MMMModal, a multimodal instruction tuned Model (LLM) specifically designed to handle multiple modalities, including images, audio, and text in a multi-turn dialogue setting. Our novel approach focuses on aligning representations from various modality encoders into a unified space. Unlike existing methods, our model effectively able to process multi-turn dialogues and incorporate multiple images or audio inputs in its responses. We provide examples demonstrating the multi-modal understanding capabilities of MMMModal.

\bibliography{neurips_2023}{}
\bibliographystyle{unsrt}

\end{document}